\documentclass[11pt]{article}

\usepackage[final]{acl}

\usepackage{times}
\usepackage{latexsym}

\usepackage[T1]{fontenc}

\usepackage[utf8]{inputenc}

\usepackage{microtype}

\usepackage{inconsolata}

\usepackage{graphicx}

\usepackage{algorithm}
\usepackage{algpseudocode}
\usepackage{amsmath,amssymb}
\usepackage{subfigure}
\usepackage{tcolorbox}
\usepackage{booktabs}
\usepackage{multirow}
\usepackage{multicol}
\usepackage{pifont}
\usepackage{bytefield}
\usepackage{tikz}
\usepackage[table]{xcolor}

\usepackage[shortlabels,inline]{enumitem}
\usepackage{tikz,lipsum}
\usepackage{ragged2e}
\usepackage[dvipsnames]{xcolor}
\usepackage{amsmath}
\usepackage{booktabs}
\usepackage{tabularray}
\usepackage{arydshln}
\usepackage{stmaryrd}
\usepackage{marvosym}
\usepackage{colortbl}
\usepackage{multicol}
\usepackage{multirow}
\usepackage{float}
\usepackage{amsfonts}
\usepackage{amssymb}

\usepackage{cleveref}
\crefname{section}{\S}{\S}
\crefname{table}{Table}{Tables}
\crefname{figure}{Fig.}{Figs.}
\crefname{algorithm}{Alg.}{}
\crefname{ALC@unique}{Line}{Lines}
\crefname{equation}{Eq.}{Eqs.}
\crefname{appendix}{App.}{Apps.}
\crefformat{section}{\S#2#1#3}

\usepackage{soul}
\definecolor{tablegray}{RGB}{223, 242, 252}

\usepackage{todonotes}

\NewDocumentCommand{\prompt}{O{} +m}{%
\begin{tcolorbox}[
    coltitle=white,
    colframe=black,
    colback=black!5!white,
    boxrule=1pt,
    enhanced jigsaw,
    breakable,
    pad at break*=2mm,
    left=2pt,
    right=2pt,
    top=2pt,
    bottom=2pt,
    fontupper=\small,
    fontlower=\small,
    title={#1}, 
]
#2 
\end{tcolorbox}
}

\title{BanglaIPA: Towards Robust Text-to-IPA Transcription with Contextual Rewriting in Bengali}

\author{
 \textbf{Jakir Hasan\textsuperscript{1}},
 \textbf{Shrestha Datta\textsuperscript{1}},
 \textbf{Md Saiful Islam\textsuperscript{1}},\\
 \textbf{Shubhashis Roy Dipta\textsuperscript{2}},
  \textbf{Ameya Debnath\textsuperscript{1}},
\\
\\
 \textsuperscript{1}Shahjalal University of Science and Technology, BD\\
 \textsuperscript{2}University of Maryland, Baltimore County, USA
\\
 \small{
   \textbf{Correspondence:} \href{mailto:jakirhasan718@gmail.com}{jakirhasan718@gmail.com}
 }
}

\begin{document}
\maketitle


\begin{abstract}
Despite its widespread use, Bengali lacks a robust automated International Phonetic Alphabet (IPA) transcription system that effectively supports both standard language and regional dialectal texts. Existing approaches struggle to handle regional variations, numerical expressions, and generalize poorly to previously unseen words. To address these limitations, we propose \textbf{\texttt{BanglaIPA}}, a novel IPA generation system that integrates a character-based vocabulary with word-level alignment. The proposed system accurately handles Bengali numerals and demonstrates strong performance across regional dialects.
\texttt{BanglaIPA} improves inference efficiency by leveraging a precomputed word-to-IPA mapping dictionary for previously observed words. The system is evaluated on the standard Bengali and six regional variations of the \texttt{DUAL-IPA} dataset. Results show that \texttt{BanglaIPA} outperforms baseline IPA transcription models by \textbf{58.4-78.7\%} and achieves an overall mean word error rate of {11.4}\%, highlighting its robustness in phonetic transcription generation for the Bengali language.\footnote{\url{https://github.com/Jak57/BanglaIPA}}

\end{abstract}

\section{Introduction}
\label{sec:introduction}

The International Phonetic Alphabet (IPA) is a widely accepted notation system for representing the phonetic structure of languages \citep{IPA1999}, providing precise pronunciation guidelines for learners, linguists, and speech-processing applications \citep{daniels1996world}. 
%
%
IPA plays a critical role in text-to-speech systems \citep{zhang2021revisiting}, enabling accurate and consistent speech synthesis across diverse languages.
%
%
However, many low-resource languages, including Bengali, still lack a standardized and efficient IPA transcription framework \citep{islam2024transcribing}, which poses a significant bottleneck for real-time speech generation and computational linguistic research.
%
%
Despite being spoken by approximately 260 million people worldwide \citep{islam2025banglalem}, Bengali continues to exhibit many inconsistencies in phonetic representation due to unresolved phonetic analyses and persistent linguistic ambiguities \citep{kamal2025bengali}.
%

%
The rich diversity of regional dialects in the Bengali language \citep{hasan2025banglatalk} makes the development of an automated IPA transcription system even more challenging. 
%
%
These dialects often diverge significantly from the standard language in terms of phonology, vocabulary, and syntactic structure \citep{hasan2024credibility}.
%
%
In addition, Bengali numerals exhibit context-dependent pronunciations, which existing IPA transcription systems fail to transcribe accurately \citep{AlZubaer2020BanglaIPA}.
%
%
The widespread use of the Bengali language highlights the need for a specialized IPA transcription framework, particularly to support high-quality text-to-speech synthesis \citep{nath2024ai}.

%

\begin{figure}[!t]
    \centering
    \begin{tikzpicture}
        \node[rounded corners=20pt, draw=none, inner sep=0pt, outer sep=0pt, clip] 
        {\includegraphics[width=\linewidth]{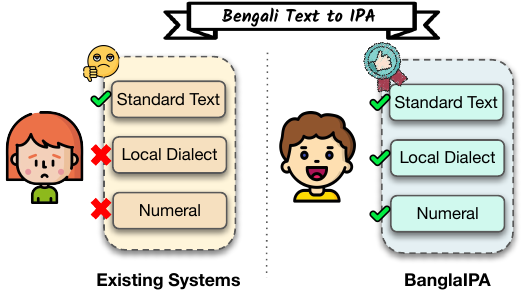}};
    \end{tikzpicture}
    \caption{Existing IPA transcription systems (left) are limited to standard Bengali and struggle to process texts containing regional dialectal variations and numerical expressions. In contrast, \texttt{BanglaIPA} (right) successfully transcribes both \textbf{standard and regional dialects} and accurately \textbf{handles numerical expressions} through the incorporation of a \textbf{contextual rewriting} procedure.}
    \label{fig:intro}
\end{figure}

In this work, we introduce \texttt{BanglaIPA}, the first end-to-end IPA transcription system designed to support standard Bengali, six regional dialects, and numerical expressions.
\texttt{BanglaIPA} integrates multiple processing modules that combine data-driven machine learning with algorithmic techniques to generate accurate and robust phonetic transcriptions of Bengali text \citep{zhou2021machine}.
As illustrated in \cref{fig:intro}, existing systems are primarily developed for standard Bengali, which severely limits their applicability for users of regional dialects.
Moreover, these systems are unable to properly transcribe numbers, instead converting individual Bengali digits into English forms\footnote{\url{https://ipa.bangla.gov.bd/}}. 
\texttt{BanglaIPA} addresses these limitations through the incorporation of a contextual rewriting \citep{bao2021contextualized} mechanism and by training the transcription generation model on both standard and regional dialectal data.
%

%
In the \texttt{BanglaIPA} system, input text containing numbers is first rewritten into word form by leveraging the full textual context.  
%
Like Bengali, in English, the digit `{1}' is pronounced differently in ``{1} dollar'' and ``{1}st place'' despite sharing the same numeric form.
%
%
The contextual rewriting ensures the preservation of the correct pronunciation of numbers.
%
%
For each word in the rewritten text, IPA transcription is generated using a Transformer-based model \citep{vaswani2017attention} trained on the \texttt{DUAL-IPA} \citep{fatema2024ipa} dataset. 
The resultant word-IPA pairs are cached in a dictionary for efficient reuse.
%
%
%
We develop a State Alignment (\texttt{STAT}) algorithm to specify which subword segments require model-based transcription, enabling robust handling of out-of-vocabulary characters and symbols.
%
%
%
Results demonstrate that our approach achieves a mean word error rate of {11.4}\% on the constructed test set, which is a substantial improvement of {58.4-78.7}\% over the baseline \texttt{MT5} \citep{xue2020mt5} and \texttt{UMT5} \citep{chung2023unimax} models.
%
%
Moreover, \texttt{BanglaIPA} maintains a strong and consistent performance across all regional dialects, with word error rates remaining close to {11}\% for four of the six regions.
%
%
%

In summary, our main contributions are: 

\begin{itemize}

    \item We propose \texttt{BanglaIPA}, the first end-to-end system capable of generating phonetic representation (IPA) of the standard Bengali language, its six regional dialects, and numbers.
    \item \texttt{BanglaIPA} substantially outperforms existing baseline models on the \texttt{DUAL-IPA} dataset and demonstrates strong performance across regional variations with low word error rates.
    \item We introduce the State Alignment (\texttt{STAT}) algorithm, which enables efficient handling of out-of-vocabulary characters and symbols.
    
\end{itemize}
%
    
    
    

\section{Methodology}
\label{sec:methodology}

\begin{figure}[!t]
    \centering
    \begin{tikzpicture}
        \node[rounded corners=0pt, draw=none, inner sep=0pt, outer sep=0pt, clip] 
            {\includegraphics[width=\linewidth]{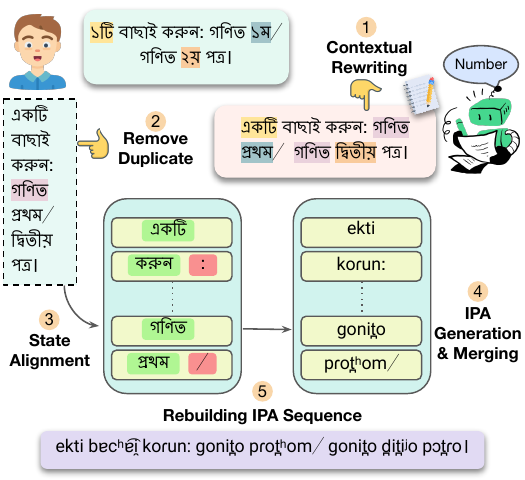}};
    \end{tikzpicture}
    \caption{Overview of the \textbf{end-to-end processing} pipeline of the \texttt{BanglaIPA} system for generating the \textbf{International Phonetic Alphabet (IPA)} transcription. 
    }
    \label{fig:banglatalk_system}
\end{figure}


%
The \textbf{\texttt{BanglaIPA}} system comprises multiple processing modules designed to accurately transcribe Bengali text into the International Phonetic Alphabet (IPA).
%
%
An overview of the complete processing pipeline is shown in \cref{fig:banglatalk_system} using an example script.

\paragraph{Contextual Rewriting} 
%
The pronunciation of Bengali numerals is highly context-dependent, and the same numeric expression may correspond to multiple IPA transcriptions, posing significant challenges for rule-based approaches.
%
%
%
Examples illustrating this pronunciation variability are provided in Appendix \ref{sec:number}.
%
%
To address this issue, we incorporate a large language model (LLM), which offers strong contextual reasoning capabilities \citep{zhu2024can}.
%
%
Specifically, we prompt the \texttt{GPT-4.1-nano} model (prompt in \cref{fig:prompt}) to rewrite only the numerical expressions in the input text into their corresponding word forms based on the surrounding textual context. 
This rewriting step ensures that numerals are converted into linguistically appropriate lexical forms before generating IPA transcription.

\begin{algorithm}[t]
\caption{The State Alignment algorithm segments a word into subwords using a predefined vocabulary. It assigns a state to each segment to indicate the need for model-based IPA generation.}
\label{alg:state_alignment}
\begin{algorithmic}[1]
   \Require token $t$, set of characters $\text{C}$
    \Ensure state and subtoken lists: $S$, $T$
   \State Initialize $S \gets []$, $T \gets []$
   \State $N \gets$ length of $t$
   \State $i \gets 0$
   \While{$i < N$}
       \State $st \gets \textbf{false}$
       \State $t_s$ $\gets \text{\textquotedblleft}\text{\textquotedblright}$
       \If{$t[i] \in \text{C}$}
           \State $st \gets \textbf{true}$
           \While{$i < N$ \textbf{and} $t[i] \in \text{C}$}
               \State Append $t[i]$ to $t_s$
               \State $i \gets i + 1$
           \EndWhile
       \Else
           \While{$i < N$ \textbf{and} $t[i] \notin \text{C}$}
               \State Append $t[i]$ to $t_s$
               \State $i \gets i + 1$
           \EndWhile
       \EndIf
       \State Append $st$ to $S$
       \State Append $t_s$ to $T$
   \EndWhile
\end{algorithmic}
\end{algorithm}
 
\paragraph{De-duplication} 
%
The contextually rewritten text is used to construct a dictionary in which each unique word serves as a key and its corresponding IPA transcription is stored as the value.
%
%
For each newly encountered word, an empty IPA entry is initially assigned and subsequently populated by downstream processing modules.
%
%
This de-duplication strategy ensures that an IPA transcription is generated only once per unique word and reused thereafter, thereby reducing redundant computations of the system. 

\paragraph{State Alignment} 
%
The unique words extracted from the rewritten text may contain characters or symbols that do not belong to Bengali language.
%
%
Such out-of-vocabulary characters and symbols present significant challenges for accurate IPA generation.
%
%
To address this issue, we propose the State Alignment (\texttt{STAT}) algorithm (see \cref{alg:state_alignment}), which leverages a predefined Bengali character set to split every word into subword segments. 
Each segment is assigned a state to represent whether it needs model-based IPA generation or has to be kept unchanged.
This mechanism ensures that any segment with foreign characters and symbols is not passed to the model that was not seen during training, enabling robust handling of unseen characters.
%
%
%


\paragraph{Generation of IPA \& Merging} 
%
For each subword identified as requiring IPA generation, we construct a synthetic sentence by inserting spaces between adjacent characters within the subword and then apply vectorization.
%
%
The resulting vectorized input is passed to a Transformer-based model, which generates sequences of IPA characters. 
%
%
After generation, the spaces between the characters are removed to produce the final IPA representation of the subword.
%
%
The complete phonetic transcription for each word is obtained by merging the model-generated IPA for the relevant subwords with the remaining segments that do not require direct transcription.
This merging process ensures a coherent and accurate word-level phonetic transcription.



\paragraph{Rebuilding Sequence} 
%
The merged IPA transcriptions are used to update the \texttt{Word-IPA} dictionary.
%
%
The contextually rewritten text is then traversed from beginning to end, and the corresponding IPA entries are retrieved from the dictionary to reconstruct the final IPA transcription of the Bengali text.

\begin{figure}[!t]
    \centering
    \includegraphics[width=1\linewidth]{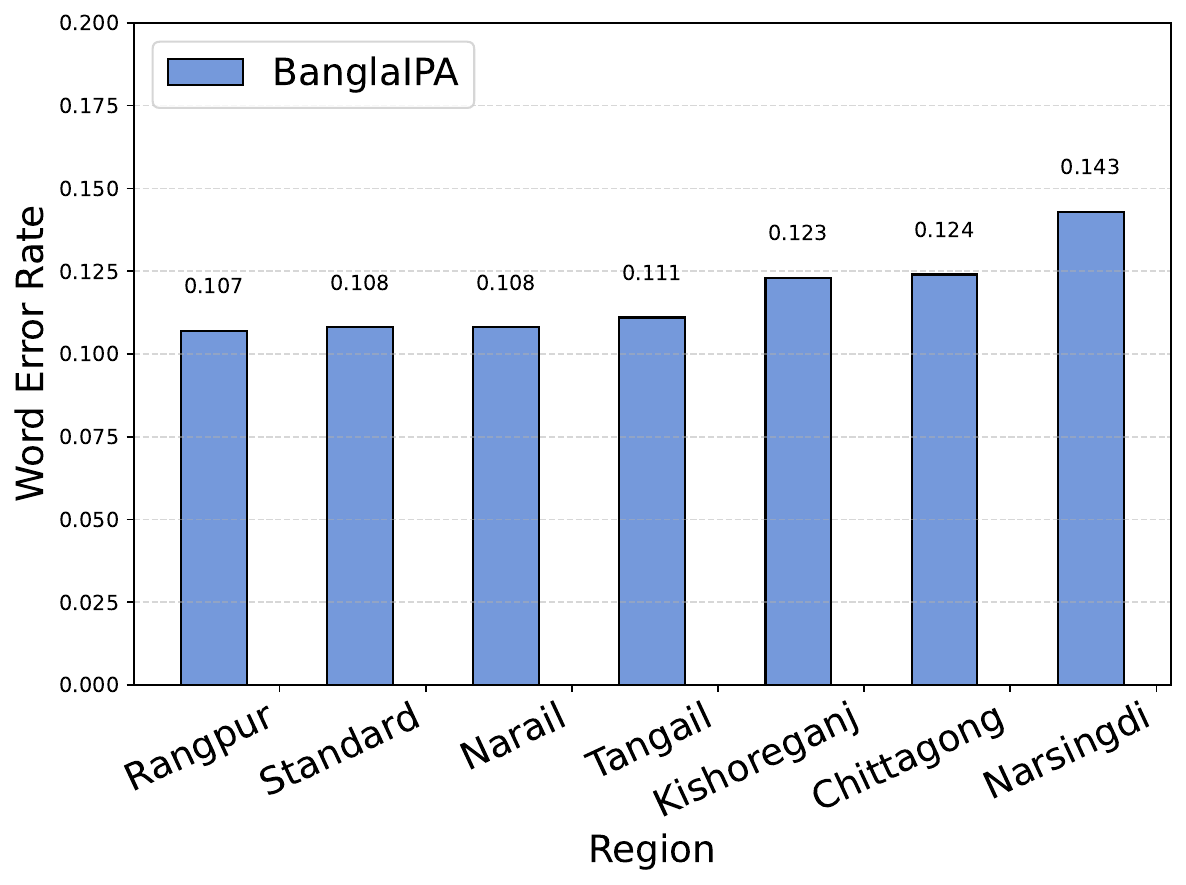}
    \caption{Region-wise word error rate distribution on the \texttt{DUAL-IPA} test dataset, with IPA transcriptions generated by the \texttt{BanglaIPA} system. The system consistently shows \textbf{very low word error rates} across most regions.}
    \label{fig:plot_wer}
\end{figure}

\begin{table*}[!t]
\centering
\resizebox{0.95\textwidth}{!}{%
\begin{tabular}{lcccccccc}
  \toprule
  \multirow{2}{*}{\textbf{Model}} & \multicolumn{7}{c}{\textbf{Word Error Rate} $\downarrow$} & \multirow{2}{*}{\textbf{Mean}}\\
  \cmidrule{2-8}
                           & Chittagong & Kishoreganj & Narail & Narsingdi & Standard & Rangpur & Tangail &  \\ 
  \midrule
  MT5      & 27.8 & 39.7 & 60.4 & 67.1 & 43.4 & 106.4 & 88.1 & 53.5 \\
  UMT5     & 31.6 & 22.8 & 19.5 & 28.5 & 28.6 & 29.0  & 27.8 & 27.4 \\
  BanglaIPA & \textbf{12.4} & \textbf{12.3} & \textbf{10.8} & \textbf{14.3} & \textbf{10.8} & \textbf{10.7} & \textbf{11.1} & \textbf{11.4} \\
  \bottomrule
\end{tabular}
}
\caption{Performance comparison of different models on the \texttt{DUAL-IPA} dataset for Bengali text-to-IPA transcription.}
\label{tab:compare_wer}
\end{table*}
\section{Results \& Discussion}
\label{sec:result_and_discussion}

\subsection{Implementation Details}
%
We train a small Transformer-based model comprising a single encoder-decoder architecture with approximately 8.6 million parameters. 
%
%
%
Model training is performed using the \texttt{RMSprop} optimizer \citep{elshamy2023improving} in conjunction with a sparse categorical cross-entropy loss function \citep{mao2023cross}.
%
%
%
The model is trained on the processed \texttt{DUAL-IPA} dataset \citep{fatema2024ipa} as described in \cref{sec:dataset} for 40 epochs, using a learning rate of {0.001} and a batch size of {64}.

\subsection{Evaluating \texttt{BanglaIPA}}

%
To evaluate the performance of our IPA transcription system, \textbf{\texttt{BanglaIPA}}, we conduct experiments on the prepared test set described in \cref{sec:dataset} from the \texttt{DUAL-IPA} dataset.
%
%
This test set includes standard Bengali as well as six regional variations spoken in Bangladesh: Chittagong, Kishoreganj, Narail, Narsingdi, Rangpur, and Tangail. 
%
%
The distribution of text samples across these regions is detailed in \cref{tab:dual_ipa}. 
%
%
We compare the performance of \texttt{BanglaIPA} against two baseline models, \texttt{MT5} \citep{xue2020mt5} and \texttt{UMT5} \citep{chung2023unimax}, using word error rate (WER) as the evaluation metric.
%
%

%
As shown in \cref{tab:compare_wer}, \texttt{BanglaIPA} consistently outperforms both baseline models across all regions, achieving an overall word error rate improvement of {58.4-78.7}\%.
%
%
Among the baselines, \texttt{MT5} exhibits the weakest performance, with a WER of {53.5}\%.
%
%
Although \texttt{UMT5} performs better, achieving a WER of {27.4}\%, it still falls substantially short of the performance of \texttt{BanglaIPA}.
%
%
The low overall WER of \textbf{11.4\%} in our system represents a significant advancement in automated IPA transcription. 

\paragraph{Performance across Regions} 

In \texttt{BanglaIPA}, dialect-specific IPA is generated for dialectal vocabulary, while standard IPA is produced for standard vocabulary.
%
We further analyze the system's regional performance variability. 
%
%
%
As illustrated in \cref{fig:plot_wer}, the word error rate remains close to {11}\% for four of the seven evaluated regions.
%
%
The lowest word error rate of {10.4}\% is observed for the Rangpur dialect.
%
%
%
%
%
Compared to the baseline systems, \texttt{BanglaIPA} demonstrates more consistent performance across Bengali regional variations. This robustness can be attributed to training the Transformer-based model from scratch, whereas the baselines are fine-tuned on the text-IPA pairs.
%

\paragraph{Numerical Rewriting Analysis}


For assessing the impact of the LLM-based contextual rewriting stage on the IPA generation pipeline, we construct a dataset of eighteen sentences containing numerical expressions in diverse linguistic contexts, along with their corresponding human-validated rewritten forms. Experimental results show that, in the absence of the contextual rewriting stage, the word error rate between the original text containing numbers and the human-validated rewritten text is 27\%. In contrast, incorporating the LLM-based rewriting stage reduces the WER to 1.3\% when compared against the same human-validated references.
This substantial reduction in error is critical for the downstream components of the \texttt{BanglaIPA} system, which operate on the rewritten text. Although the contextual rewriting stage is not perfectly accurate and may occasionally produce hallucinated outputs, its overall performance represents a significant improvement over the non-rewriting baseline and is sufficient for practical use within the system.

\section{Related Work}



\subsection{Text-to-IPA in Bengali} \label{sec:rel_bengali}
%
In recent years, several International Phonetic Alphabet (IPA) transcription systems have been proposed for the Bengali language.
%
%
A rule-based approach introduced by \citet{AlZubaer2020BanglaIPA} focuses on transcribing standard Bengali characters, words, and numerals.
%
%
%
The District Guided Token (\texttt{DGT}) method proposed by \citet{islam2024transcribing} is designed to handle only regional dialects by finetuning \texttt{T5}-based models \citep{raffel2020exploring} with district-level information.
%
%
The challenges inherent to Bengali IPA transcription are comprehensively analyzed by \citet{fatema2024ipa}, who also introduced a novel transcription framework along with the \texttt{DUAL-IPA} dataset, which serves as a benchmark for evaluating transcription systems for Bengali.

\subsection{Text-to-IPA in Foreign Languages} \label{sec:rel_other}
%
High-resource languages such as English \citep{engelhart2021grapheme}, Mandarin \citep{odinye2015phonology}, German \citep{odom2023german}, and French have seen substantial advances in automatic IPA transcription in recent years.
%
%
In particular, \citet{yolchuyeva2020transformer} investigates the effectiveness of attention-based Transformer architectures for the grapheme-to-phoneme (G2P) conversion task \citep{deri2016grapheme}, demonstrating notable improvements over earlier convolutional approaches \citep{yolchuyeva2019grapheme}.
%
%
More broadly, \citet{cheng2024survey} presents a comprehensive survey of neural network-based methods for grapheme-to-phoneme conversion in both monolingual and multilingual settings, outlining current challenges and promising directions.

\section{Conclusion \& Future Work}
\label{sec:conclusion}

%
In this work, we present \texttt{BanglaIPA}, the first end-to-end system designed to generate International Phonetic Alphabet (IPA) transcriptions of standard Bengali, six regional dialects, and numerals.
%
%
\texttt{BanglaIPA} employs a contextual rewriting mechanism to handle context-dependent numerical pronunciations effectively.
%
%
IPA transcriptions are generated on a word-by-word basis using a lightweight Transformer architecture trained on the \texttt{DUAL-IPA} dataset. 
%
%
To address out-of-vocabulary characters and symbols, we introduce the State Alignment (\texttt{STAT}) algorithm that applies model-based transcription generation only to valid subwords.
%
%
%
Experimental results demonstrate that the proposed system outperforms baseline models, achieving a minimum word error rate of {11.4}\%.
%
%
As future work, we plan to incorporate data from additional regional dialects and train a small language model for efficient contextual rewriting of numbers.

\section*{Limitations}
The proposed system is trained on data covering six major regional dialects of Bengali. While this provides broad linguistic coverage, performance on dialects that are not represented in the training data may vary, and extending coverage to additional dialects remains an important direction for future work. In addition, the system employs an LLM-based module for context-aware rewriting of numerical expressions. Although this component improves robustness and consistency in numerical reasoning, it introduces modest additional computational cost, which we view as an acceptable trade-off given the gains in accuracy and generalization.
    




\bibliography{custom}

\appendix

\newpage
\section*{Appendix}

\section{\texttt{DUAL-IPA} Dataset Detail}
\label{sec:dataset}

\begin{algorithm}[!t]
\caption{\textsc{SplitDataset}: Construction of training and test sets from the \texttt{DUAL-IPA} dataset using the IPA novelty score.}
\label{alg:SplitDataset}
\begin{algorithmic}[1]
\Require Dataset $D$, Elbow point score $EP$
\Ensure Training set $D_{\text{train}}$, Test set $D_{\text{test}}$
\State Initialize $D_{\text{test}} \gets \emptyset$
\State Initialize $D_{\text{train}} \gets D$
\State $CS \gets \infty$
\While{$CS \geq 0$}
    \For{each sample $S$ in $D_{\text{train}}$}
        \State Extract word IPA set $S_S$ from $S$
        \State Extract word IPA set $S_T$ from $D_{\text{test}}$
        \State $CS(S) \gets |S_S - S_T|$ 
    \EndFor
    \State Select $S^{*} \gets \arg\max_S CS(S)$
    \State $CS \gets CS(S^{*})$
    \If{$CS < EP$}
        \State \textbf{break}
    \EndIf
    \State Move $S^{*}$ from $D_{\text{train}}$ to $D_{\text{test}}$
\EndWhile
\State \Return $D_{\text{train}}, D_{\text{test}}$
\end{algorithmic}
\end{algorithm}

The \texttt{DUAL-IPA} dataset \citep{fatema2024ipa} represents the first effort to standardize Bengali IPA transcription by providing parallel pairs of text and corresponding IPA annotations. Two subsets of this comprehensive dataset have been made publicly available through affiliated competitions\footnote{\url{https://www.bengali.ai/}}: the \texttt{DataVerse Challenge – ITVerse 2023} \citep{dataverse_2023}, which contains a corpus of standard Bengali text, and \texttt{Bhashamul: Bengali Regional IPA Transcription} \citep{regipa}, which includes corpora from six regional dialects. In this study, we utilize these publicly available subsets to train and evaluate our novel automatic IPA transcription system.


\paragraph{Preprocessing}
In Bengali, the same character can be represented using multiple Unicode encodings \citep{needleman2000unicode}, which introduces ambiguity into the text \citep{ansary2023unicode}. Since neural models typically perform better when trained on standardized and normalized inputs \citep{demir2022graph}, we apply a Bengali text normalization procedure using the normalizer proposed by \citet{hasan-etal-2020-low}. This preprocessing step ensures consistent character representations and improves the reliability of downstream modeling.

\begin{table}[!t]
\centering
\begin{tabular}{@{}lccc@{}}
\toprule
\multirow{2}{*}{\textbf{Region}} & \multicolumn{3}{c}{\textbf{Number of Sample}} \\
\cmidrule{2-4}
& Train & Test & Total\\
\cmidrule{1-4}
Chittagong           & 4157  & 605  & 4762  \\
Kishoreganj          & 6089  & 642  & 6731  \\
Narail               & 5449  & 573  & 6022  \\
Narsingdi            & 3923  & 586  & 4509  \\
Rangpur              & 3539  & 503  & 4042  \\ 
Tangail              & 3732  & 513  & 4245  \\
Standard             & 18882 & 3117 & 21999 \\
\midrule
Total                & 45771 & 6539 & 52310 \\
\bottomrule
\end{tabular}
\caption{Distribution of the processed training and testing data in the \texttt{DUAL-IPA} dataset of parallel text corpus across the standard Bengali and its six regional dialects.}
\label{tab:dual_ipa}
\end{table}



\paragraph{Elbow Method for Dataset Splitting}
We use the normalized dataset to construct the training and test sets, as the original test split of the \texttt{DUAL-IPA} dataset is not publicly available. To minimize lexical overlap between the two splits and to maximize the presence of unseen words in the test set, we employ a custom dataset splitting procedure, described in \cref{alg:SplitDataset}.
At each iteration, the sample with the highest IPA novelty score is selected and assigned to the test set. This novelty score is computed based on the number of words in the sample that have not yet appeared in the test set. The test set construction continues until the novelty score falls below a predefined Elbow Point (\texttt{EP}) \citep{liu2020determine}, after which the remaining samples are allocated to the training set.

\begin{figure}[!t]
    \centering
    \includegraphics[width=1\linewidth]{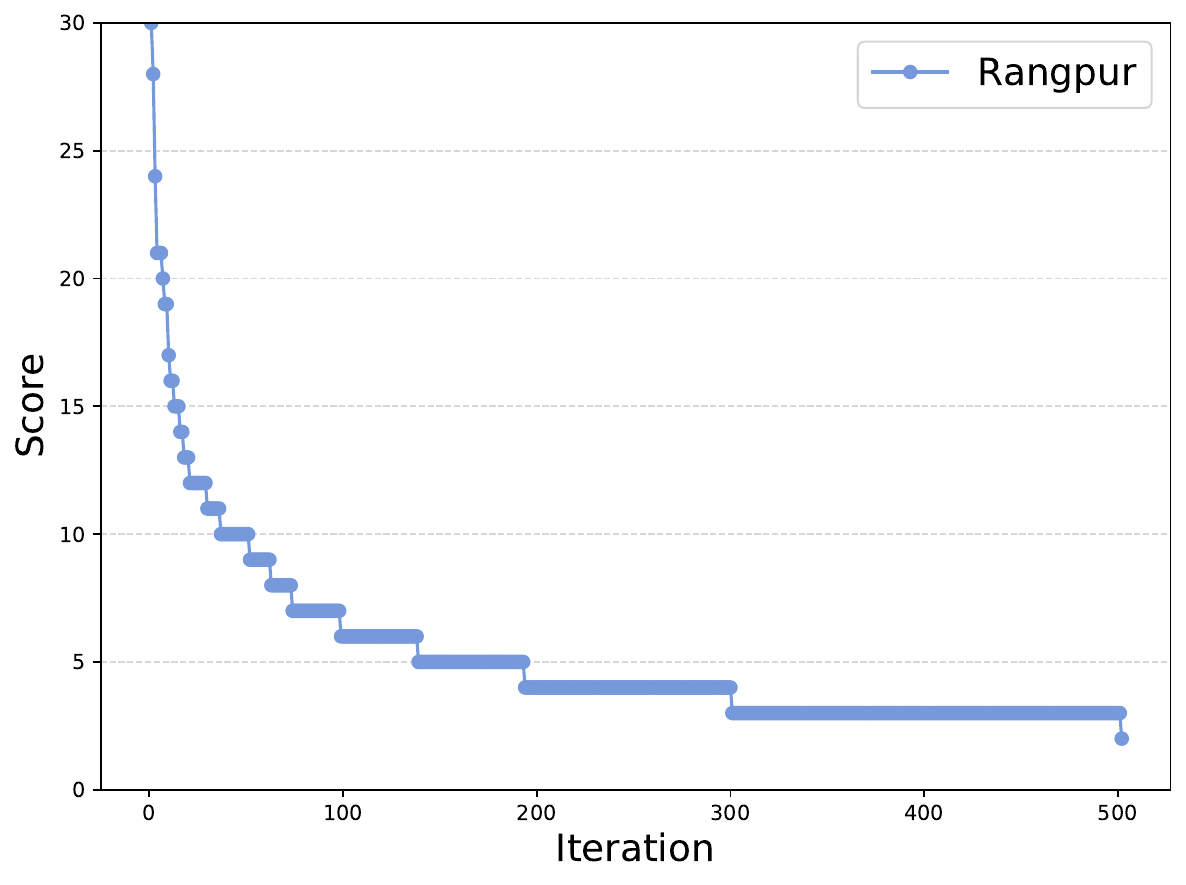}
    \caption{Identification of the optimal Elbow Point (\texttt{EP}) to terminate the dataset splitting procedure for the Rangpur dialect using the Elbow method.}
    \label{fig:plot_elbow}
\end{figure}

\cref{fig:plot_elbow} presents the novelty score progression across iterations for the Rangpur dialect. The Elbow Point (\texttt{EP}) is identified when the curve flattens and becomes approximately parallel to the x-axis, indicating that further iterations result in negligible changes to the score. In our experiments, the \texttt{EP} is observed at a score of approximately 3.
Based on this criterion, we construct the training and test splits shown in \cref{tab:dual_ipa}. 
%
%
The resulting test set maximizes the number of unseen words and lexical diversity \citep{malvern2004lexical, johansson2008lexical}.

\section{Pronunciation Change of Number}
\label{sec:number} 

\begin{figure}[!t]
    \centering
    \includegraphics[width=1\linewidth]{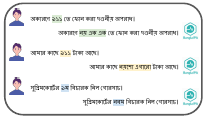}
    \caption{The pronunciation of Bengali numbers varies depending on the textual context (highlighted in colors). The \texttt{BanglaIPA} system converts numbers into word forms as an intermediate step in IPA transcription.}
    \label{fig:plot_num}
\end{figure}


Numbers are a fundamental component of any language, conveying critical information in everyday communication \citep{wiese2003numbers}. Accurate interpretation of numerical expressions is essential for a robust IPA transcription system in Bengali. Numerical expressions representing quantities, years, dates, times, phone numbers, and similar constructs can have multiple verbal pronunciations and word forms, as illustrated in \cref{fig:plot_num}.
In the \texttt{BanglaIPA} system, we address this challenge by converting numbers into intermediate word-form representations using \texttt{GPT-4.1-nano}. We provide the model with the prompt shown in \cref{fig:prompt}, instructing it to leverage the contextual information of the full sentence \citep{an2024make} to generate word forms for numerical expressions while leaving other words unchanged. The resulting contextually enriched text is subsequently used for model-based IPA transcription, improving the accuracy and robustness. 






\begin{center}
\begin{minipage}{\columnwidth} 
\begin{tcolorbox}[colback=gray!5!white,
                  colframe=gray!50!black,
                  colbacktitle=gray!75!black]

   \textbf{System Prompt: } You are a helpful chatbot who understands Bengali numerals in different contexts.\\[6pt]
   \textbf{User Prompt: } Please rewrite the provided text so that no Bengali digits are present. Convert the numbers to word form based on the context. Do not modify any words.\\[6pt]
   Here is the text: $\texttt{\{user\_text\}}$.

\end{tcolorbox}
\captionof{figure}{Prompt used to instruct the large language model to generate contextually rewritten Bengali text by converting numerical expressions into word forms.}
\label{fig:prompt}
\end{minipage}
\end{center}





\label{sec:appendix}

\end{document}